\documentclass[conference]{IEEEtran}
\IEEEoverridecommandlockouts
\usepackage{cite}
\usepackage{amsmath,amssymb,amsfonts}
\usepackage{algorithmic}
\usepackage{graphicx}
\usepackage{textcomp}
\usepackage{xcolor}
\usepackage{pifont}
\usepackage{float}
\usepackage{url}
\usepackage{graphicx}
\usepackage{subcaption}

\usepackage{hyperref}
\hypersetup{
    colorlinks=true,
    allcolors=blue
}

\usepackage{afterpage}
\usepackage{booktabs}
\usepackage{tabularray}
\usepackage{microtype}
\usepackage{multirow}
\usepackage{array}
\newcommand{\PreserveBackslash}[1]{\let\temp=\\#1\let\\=\temp}
\newcolumntype{C}[1]{>{\PreserveBackslash\centering}p{#1}}
\newcolumntype{R}[1]{>{\PreserveBackslash\raggedleft}p{#1}}
\newcolumntype{L}[1]{>{\PreserveBackslash\raggedright}p{#1}}
\usepackage{booktabs}

\def\BibTeX{{\rm B\kern-.05em{\sc i\kern-.025em b}\kern-.08em
    T\kern-.1667em\lower.7ex\hbox{E}\kern-.125emX}}
\begin{document}

\title{MangoLeafViT: Leveraging Lightweight Vision Transformer with Runtime Augmentation for Efficient Mango Leaf Disease Classification}


\author{ 
    \IEEEauthorblockN{
    Rafi Hassan Chowdhury, 
    Sabbir Ahmed\\}

    \IEEEauthorblockA{
        Department of Computer Science and Engineering, Islamic University of Technology, Gazipur 1704, Bangladesh}

    \IEEEauthorblockA{\{rafihassan, sabbirahmed\}@iut-dhaka.edu
    } 
}


\makeatletter
\let\old@ps@IEEEtitlepagestyle\ps@IEEEtitlepagestyle
\def\confheader#1{%
    \def\ps@IEEEtitlepagestyle{%
        \old@ps@IEEEtitlepagestyle%
        \def\@oddhead{\strut\hfill#1\hfill\strut}%
        \def\@evenhead{\strut\hfill#1\hfill\strut}%
    }%
    \ps@headings%
}
\makeatother
\confheader{
        \parbox{20cm}{2024 27th International Conference on Computer and Information Technology (ICCIT)\\
        20-22 December 2024, Cox’s Bazar, Bangladesh}
}

\IEEEpubid{
\begin{minipage}[t]{\textwidth}\ \\[10pt]
      \small{Accepted in 27th ICCIT \copyright2024 IEEE }  
\end{minipage}
}

\maketitle

\begin{abstract}
Ensuring food safety is critical due to its profound impact on public health, economic stability, and global supply chains. Cultivation of Mango, a major agricultural product in several South Asian countries, faces high financial losses due to different diseases, affecting various aspects of the entire supply chain. While deep learning-based methods have been explored for mango leaf disease classification, there remains a gap in designing solutions that are computationally efficient and compatible with low-end devices. In this work, we propose a lightweight Vision Transformer-based pipeline with a self-attention mechanism to classify mango leaf diseases, achieving state-of-the-art performance with minimal computational overhead. Our approach leverages global attention to capture intricate patterns among disease types and incorporates runtime augmentation for enhanced performance. Evaluation on the MangoLeafBD dataset demonstrates a 99.43\% accuracy, outperforming existing methods in terms of model size, parameter count, and FLOPs count. 

\end{abstract}

\begin{IEEEkeywords}
Lightweight models, MobileViT, Smart Agriculture, Transfer Learning, Food Safety, Attention Mechanism
\end{IEEEkeywords}

\section{Introduction}

Mango is a globally significant tropical fruit, prized for its rich flavor and nutritional value, being a key source of antioxidants, dietary fiber, and vitamins. Beyond its culinary appeal, mango plays a crucial role in the agricultural economies of many tropical and subtropical regions. In 2023, global mango production reached 78.56 billion kilograms, with imports valued at 3.57 billion USD and exports totaling 2.34 billion USD. India remains the top producer, while Bangladesh ranks eighth, contributing 1.5 billion kilograms to global production \cite{Tridge2023}.

Traditional methods for disease classification, reliant on manual expert evaluation, are inefficient, time-consuming, and often inaccessible in rural areas, leading to delayed detection and substantial economic losses due to reduced crop yields \cite{ahmed2022less}.  This underscores the need for automated, accurate, and efficient disease detection systems \cite{hughes2015open, li2021plant, rafi2023criticalAnalysis}. Advances in Deep Learning have facilitated the development of such systems with Vision Transformer (ViT) models demonstrating exceptional performance in various image classification tasks, including plant disease detection \cite{THAKUR2023102245, khan2022rethinking, thakur2022vision}.

While Machine Learning (ML) and Deep Learning (DL) models have been widely explored for leaf disease classification \cite{ahmed2024ExE,  ashmafee2023anEfficient, kamilaris2018deep, herok2023cotton}
there has been limited focus on mango leaf disease classification. This is due to several challenges, including the lack of large datasets, resource constraints, and the subtle visual similarities between different diseases \cite{Varma2024,10585939}. Ahmed \textit{et al.}  addressed this gap by curating the `MangoLeafBD' dataset, which contains a substantial number of samples across seven disease categories from diverse regions of Bangladesh \cite{AHMED2023108941}.

Recent efforts to classify mango leaf diseases have yielded promising results. Mahbub \textit{et al.} \cite{10101648} achieved 98\% accuracy using a lightweight CNN, while Rizvee \textit{et al.} \cite{RIZVEE2023100787} achieved 99.55\% accuracy with a custom CNN but at the expense of a higher FLOP count and model size, making it unsuitable for resource-constrained environments. Similarly, Swapno \textit{et al.} \cite{10585939} reported a 96.87\% accuracy using EfficientNetv2-Large, and Varma \textit{et al.} \cite{Varma2024} achieved 99.87\% accuracy using a model with 20.58 million parameters, which, due to its size, limits practical deployment on low-end devices.


Many existing models, while achieving high accuracy, are hindered by large parameter counts, model sizes, and high FLOPs, reducing their efficiency in real-world applications, particularly on lightweight devices. Recently, lightweight Vision Transformer architectures, such as MobileViT \cite{mehta2022mobilevitlightweightgeneralpurposemobilefriendly}, EfficientViT \cite{cai2024efficientvitmultiscalelinearattention}, and TinyViT \cite{wu2022tinyvitfastpretrainingdistillation}, have emerged as promising solutions. These models, leveraging transformer-based self-attention mechanisms, excel at capturing positional information, often lost in traditional convolutional architectures, while maintaining computational efficiency, making them ideal for resource-constrained environments.

In this work, we present MangoLeafViT, a lightweight Vision Transformer-based pipeline for automatic mango leaf disease classification. Leveraging the `MangoLeafBD' dataset, our approach is designed to provide high classification accuracy while maintaining low computational costs, making it suitable for deployment on low-end devices. We evaluate three lightweight Vision Transformer models: EfficientViT\_b0, TinyViT\_5m, and MobileViT\_s, and employ diverse runtime augmentation strategies to enhance generalization across disease variations. Our results demonstrate that the proposed pipeline achieves superior accuracy with significantly reduced model size, parameter count, and FLOPs, offering an efficient solution for real-world mango disease detection.

\begin{figure*}[t]
    \centering
    \begin{tabular}{cccccccc}
        \subfloat[Anthracnose]{\includegraphics[width=0.12\textwidth]{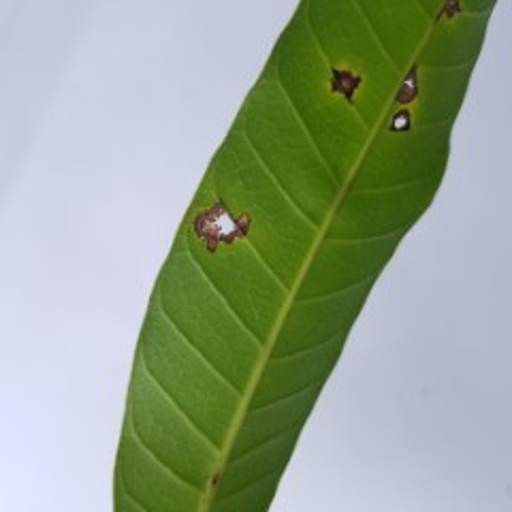}} &
        \hspace{-0.40cm} 
        \subfloat[Bacterial Canker]{\includegraphics[width=0.12\textwidth]{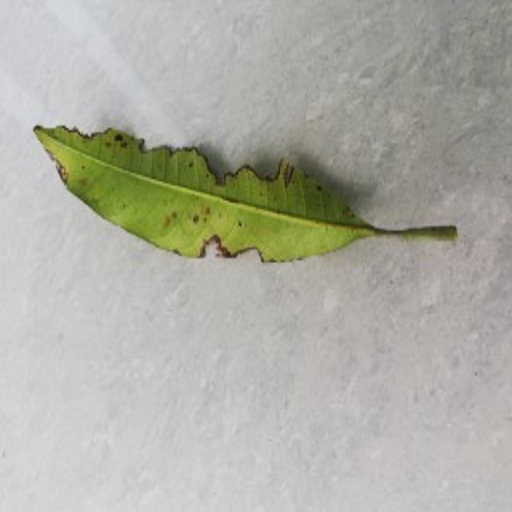}} &
        \hspace{-0.40cm} 
        \subfloat[Cutting Weevil]{\includegraphics[width=0.12\textwidth]{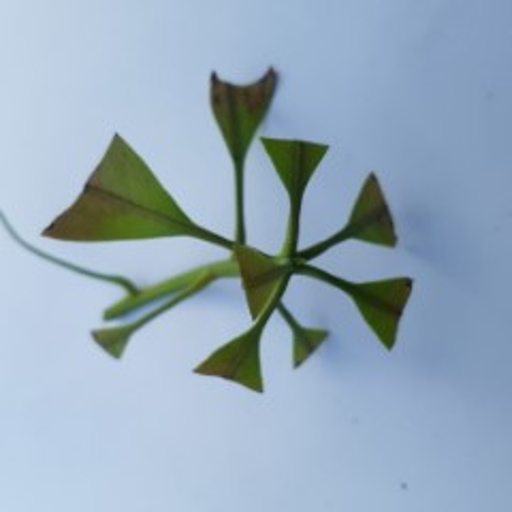}} &
        \hspace{-0.40cm} 
        \subfloat[Die Back]{\includegraphics[width=0.12\textwidth]{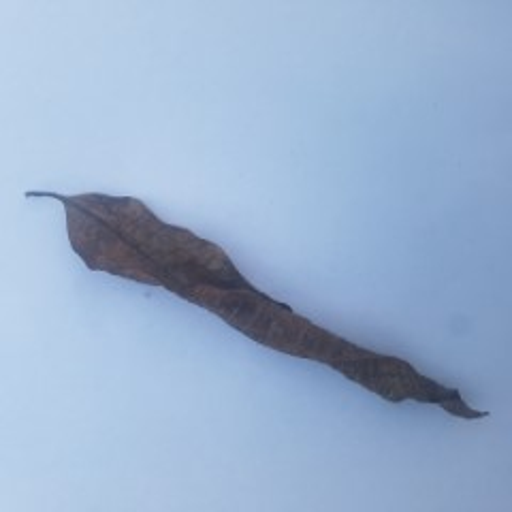}} &
        \hspace{-0.40cm} 
        \subfloat[Gall Midge]{\includegraphics[width=0.12\textwidth]{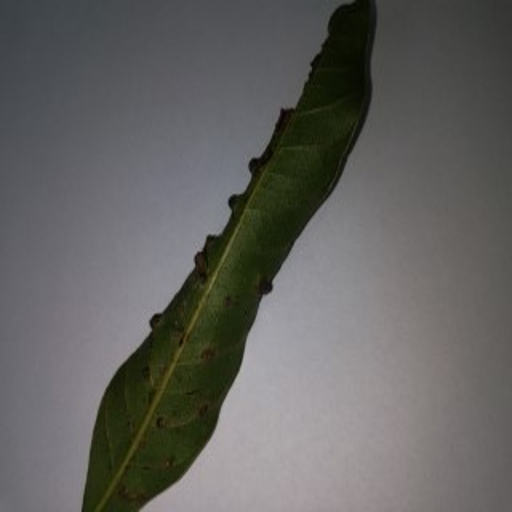}} &
        \hspace{-0.40cm} 
        \subfloat[Healthy]{\includegraphics[width=0.12\textwidth]{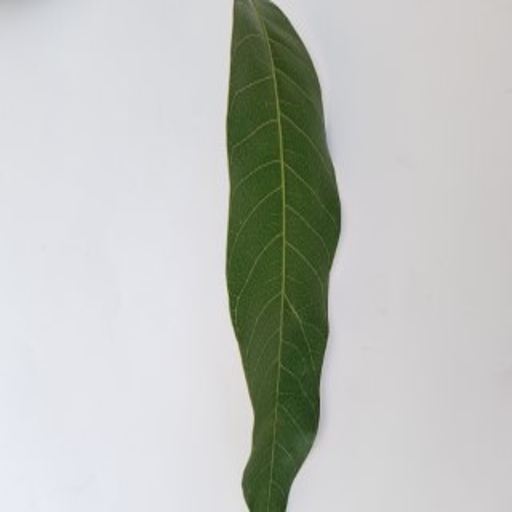}} &
        \hspace{-0.40cm} 
        \subfloat[Powdery Mildew]{\includegraphics[width=0.12\textwidth]{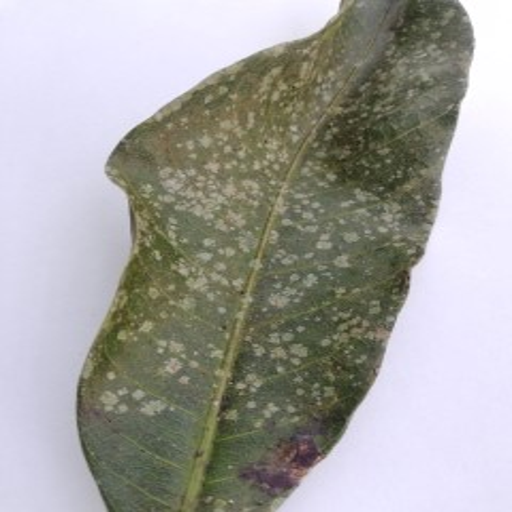}} &
        \hspace{-0.40cm} 
        \subfloat[Sooty Mould]{\includegraphics[width=0.12\textwidth]{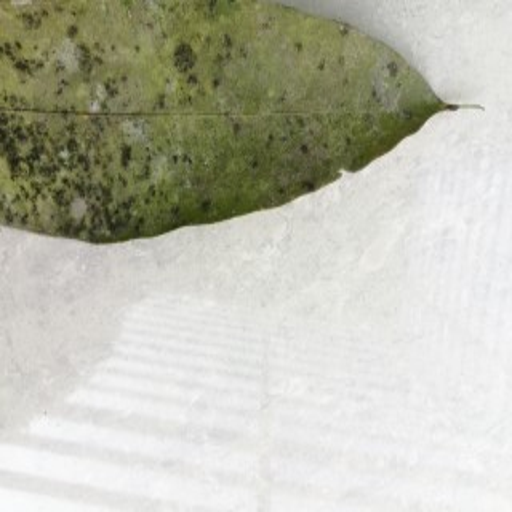}} \\
    \end{tabular}
    \caption{Sample mango leaf images of the different classes from the MangoLeafBD dataset \cite{AHMED2023108941}}
    \label{fig:overview_of_dataset_figure}
\end{figure*}

\section{Methodology} \label{Methodology}


The proposed framework takes mango leaf samples as input and outputs corresponding disease labels. The pipeline begins with a preprocessing phase, followed by model inference using pretrained architectures. We employed lightweight Vision Transformers fine-tuned on the MangoLeafBD dataset. 


\subsection{Dataset}

The MangoLeafBD dataset includes seven disease categories and a healthy class, each having 500 images \cite{AHMED2023108941}. Figure \ref{fig:overview_of_dataset_figure} presents representative samples from the eight classes. Powdery Mildew and Sooty Mould diseases have similar patterns in terms of color and disease spot. Anthracnose disease has some intra-class dissimilarity in terms of disease spot, pattern, and leaf color. The pattern of Cutting Weevil disease is different from others because this disease cuts the leaf into pieces. The images consist of close-up shots of mango leaves against a white background, providing clear visibility of the disease patterns. This dataset offers sufficient variation in features, making it well-suited for Vision Transformer-based models to effectively capture region-specific disease characteristics.

For model training and evaluation, we adopted a 5-fold cross-validation strategy, splitting the dataset into 60\% for training, 20\% for testing, and 20\% for validation. Each fold is independent, ensuring that no images are shared across the training, validation, or test sets, effectively reducing the risk of overfitting and improving the model's generalization capability.

\subsection{Data Preprocessing}
A variety of preprocessing and augmentation techniques were applied to the input images, following standard practices for enhancing model generalization during training. All augmentations were implemented at runtime, enabling the model to learn robust feature representations from diverse transformations without artificially inflating the dataset size \cite{ahmed2022less}. Samples were resized to a uniform resolution of $224\times224$, satisfying the input requirements of the ViT models used in this work.

\begin{figure}[t]
    \centering
    \begin{tabular}{cccccccc}
        \subfloat[CLAHE]{\includegraphics[width=0.15\textwidth]{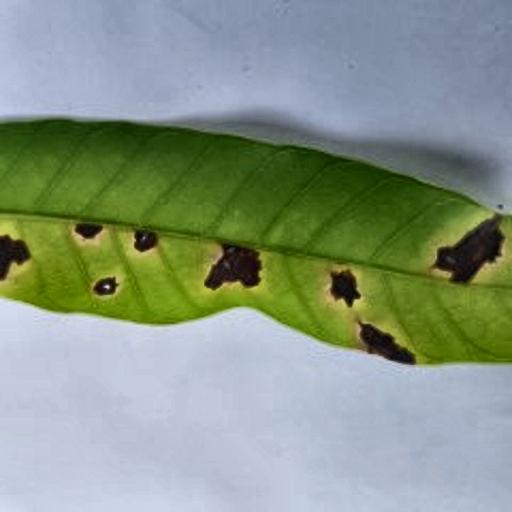}} &
        \hspace{-0.40cm} 
        \subfloat[Affine]{\includegraphics[width=0.15\textwidth]{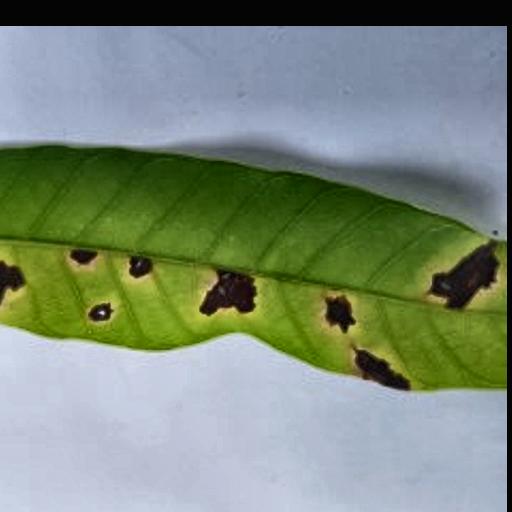}} &
        \hspace{-0.40cm} 
        \subfloat[Rotation]{\includegraphics[width=0.15\textwidth]{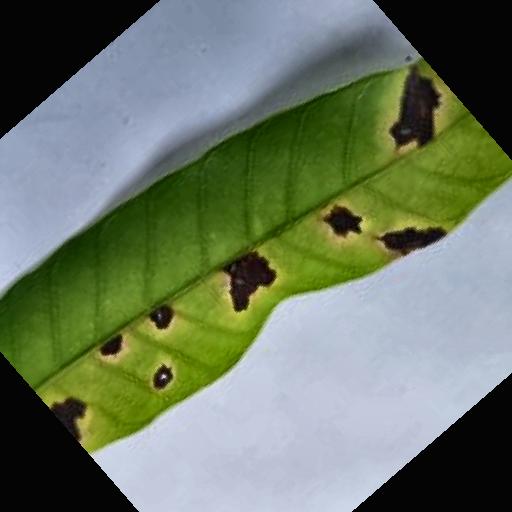}} &
        \hspace{-0.40cm} 
        \\
        \subfloat[Horizontal Flip]{\includegraphics[width=0.15\textwidth]{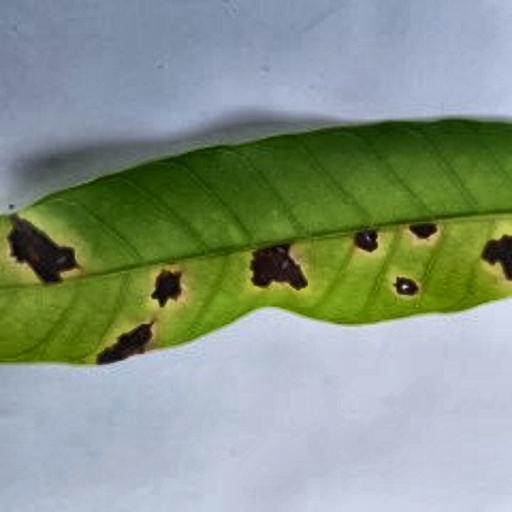}} &
        \hspace{-0.40cm} 
        \subfloat[Vertical Flip]{\includegraphics[width=0.15\textwidth]{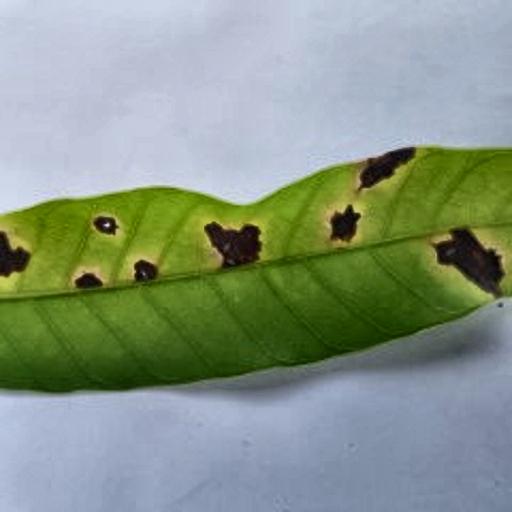}} &
        \hspace{-0.40cm} 
        \subfloat[Grayscale]{\includegraphics[width=0.15\textwidth]{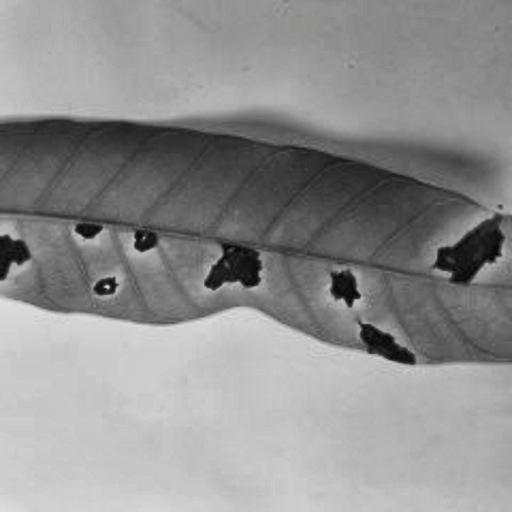}} &
        \hspace{-0.40cm} 
         \\
    \end{tabular}
    \caption{Sample augmentations performed on the images during
training, validation, and testing phase.}
    \label{fig:data_preprocessing_samples}
\end{figure}

To enhance image contrast, we applied Contrast Limited Adaptive Histogram Equalization (CLAHE) (Figure \ref{fig:data_preprocessing_samples}a), which divides the image into $8\times8$ regions, clips histogram values at a threshold of 2.0 to prevent noise amplification, and smooths blocking artifacts via bilinear interpolation. This method ensures that fine-grained texture details are retained without introducing excessive noise. 


Random rotation of up to 45 degrees was employed to account for varying orientations of mango leaves (Figure \ref{fig:data_preprocessing_samples}c). We also applied height and width shifting in both horizontal and vertical directions by a randomly selected constant factor in the range [0, 0.1]. This technique aids the model in learning spatial invariances (Figure \ref{fig:data_preprocessing_samples}b).
To further increase variability, horizontal and vertical flips were introduced, each with a probability of 50\%. These augmentations help the model learn from different viewpoints, improving robustness against orientation changes (Figures \ref{fig:data_preprocessing_samples}d and \ref{fig:data_preprocessing_samples}e). Additionally, grayscale conversion was applied with a probability of 10\% (Figure \ref{fig:data_preprocessing_samples}f), enabling the model to focus on texture and shape features, rather than relying solely on color information.

Unlike traditional offline augmentation techniques, these transformations were applied dynamically during training to prevent dataset size inflation. This runtime augmentation ensures that different variations of the same image are seen across multiple epochs, thus reducing the risk of overfitting. Moreover, this approach guarantees that no identical image appears across the training, testing, or validation splits, further improving the model’s generalization capabilities.

\subsection{Model Description}
The Vision Transformer (ViT) architecture \cite{vaswani2023attentionneed} employs a multi-headed self-attention mechanism and has recently been adapted for a wide variety of computer vision tasks. In contrast to traditional CNNs that operate on image grids, ViT divides an image into fixed-size patches, analogous to splitting text into tokens. These patches are then converted into feature vectors and processed by the transformer, which leverages feed-forward layers and self-attention to capture global relationships across the image \cite{hasan2023GaitGCN}. Positional encodings are incorporated to preserve spatial information. A classification token is appended to the sequence, and the final output of this token is used for classification tasks \cite{dosovitskiy2021imageworth16x16words}. We explored three lightweight ViT variants: MobileViT \cite{mehta2022mobilevitlightweightgeneralpurposemobilefriendly}, EfficientViT \cite{cai2024efficientvitmultiscalelinearattention}, and TinyViT \cite{wu2022tinyvitfastpretrainingdistillation}.


A lighter version of EfficientViT, known as EfficientViT\_b0 has been used in our experiment. This model is optimized for fast processing with computational requirements. The model processes images by embedding slightly overlapping patches, allowing it to extract crucial features early in the network. Its architecture follows a `Sandwich Layout', where self-attention layers are interleaved between feed-forward layers, optimizing both memory usage and computational demands. EfficientViT employs cascaded group attention and a hierarchical structure, which further enhances resource management, making it well-suited for deployment on low-end devices \cite{cai2024efficientvitmultiscalelinearattention}.


In addition to EfficientViT, we evaluated TinyViT\_5m, a member of the TinyViT series, which is designed for scenarios requiring real-time processing with limited resources. TinyViT adopts a hierarchical transformer architecture, scaled down for speed and efficiency. It utilizes window-based self-attention within its transformer blocks and depthwise convolutions between attention and multi-layer perceptron (MLP) layers, significantly reducing computational load while maintaining accuracy \cite{wu2022tinyvitfastpretrainingdistillation}. The compact design and efficiency of TinyViT\_5m make it ideal for mobile and edge device applications.


We also experimented with MobileViT\_s \cite{mehta2022mobilevitlightweightgeneralpurposemobilefriendly}, which is optimized for edge devices. MobileViT combines the strengths of CNNs for local feature extraction with the global context modeling of transformers. The architecture begins with a convolutional layer, followed by MobileNetV2 blocks \cite{sandler2019mobilenetv2invertedresidualslinear} for feature extraction, and transitions to MobileViT blocks to jointly learn local and global patterns. The use of depthwise separable convolutions and lightweight transformer layers ensures a balance between high accuracy and low computational cost. 


\subsection{Transfer Learning}
Transfer learning is a powerful approach that leverages a model pre-trained on a substantial dataset to enhance performance on a related task \cite{9869842}. This technique proves invaluable, particularly when large datasets are scarce. By utilizing a pretrained model, we can not only improve the performance but also significantly reduce training time \cite{morshed2022Fruit}.

Given that ViTs exhibit a deficiency in inductive biases typically found in CNNs, transfer learning becomes essential for effectively training on smaller datasets. Numerous studies have demonstrated that employing ViTs pretrained on expansive datasets, such as ImageNet-21k \cite{5206848}, followed by fine-tuning on a smaller dataset, can yield remarkable results that surpass many state-of-the-art benchmarks in image classification \cite{raghu2022visiontransformerslikeconvolutional}. In our experiments, we utilized ViT models pretrained on the ImageNet-1k dataset, selected for their lightweight architecture and efficiency in resource-constrained environments.


\subsection{Experimental Setup}
All experiments were conducted on a desktop system with an Nvidia RTX 3060 GPU featuring 12 GB of VRAM, paired with an Intel Core i5-13500 CPU operating at 2.50 GHz, and supported by 32 GB of RAM. The training of all models was performed in a Python environment utilizing the PyTorch library, specifically leveraging models from the `timm repository' that were pretrained on the ImageNet-1k dataset.

For the optimization of model performance, the Cross-Entropy loss function was employed, as it is well-suited for multi-class classification tasks. The training was set to a maximum of 100 epochs. Implementing k-fold cross-validation, we adopted an early stopping mechanism that halted training if the average validation loss of the current epoch exceeded the best validation loss recorded from previous epochs. This early stopping criterion was triggered if three consecutive epochs exhibited this behavior, thereby mitigating the risk of overfitting. Most models converged within 20 to 30 epochs. The learning rate was set to 0.00001, and the batch size was established at 32. The Adam optimizer was selected for this study due to its effectiveness across various architectures and its ability to facilitate rapid learning of salient features \cite{kingma2017adammethodstochasticoptimization}. 



\begin{table}[b]
\centering
\caption{Performance comparison using the baseline pretrained architectures}
\label{table:performance_analysis_of_baseline_models}
\begin{tabular}{l C{1.5cm} C{1.5cm} C{1.5cm} }
\toprule
\textbf{Architecture} & \textbf{Accuracy (\%)} & \textbf{Trainable Parameters Count (M)} & \textbf{Model Size (MB)} \\ 
\midrule


VGG16 & 90.67\% & 134.29 & 537.2 \\ 
MobileNetv2 & 91.23\% & 2.23 & 2.13 \\ 
ResNet50 & 94.77\% & 23.52 & 22.43 \\ 
ResNet34 & 95.35\% & 21.29 & 20.30 \\ 
DenseNet121 & 98.85\% & 6.96 & 6.64 \\ 
AlexNet & 99.25\% & 58.32 & 466.6 \\ 

EfficientViT\_b0 & 98.80\% & \textbf{2.14} & \textbf{2.04} \\ 
TinyViT\_5m & 99.10\% & 5.07 & 4.84 \\ 
MobileViT\_s & \textbf{99.43}\% & 4.94 & 4.71 \\ 
\bottomrule
\end{tabular}
\end{table}

\section{Results and Discussion}\label{Results and Discussion}

\subsection{Performance of different baseline architectures}
Convolutional Neural Networks (CNNs) have long been the standard in various image classification tasks. However, the recent emergence of transformer-based models has demonstrated superior performance in many scenarios, often surpassing traditional CNN architectures. To establish our baseline models, we evaluated several state-of-the-art pretrained deep CNNs alongside our vision transformer models, utilizing the `MangoLeafBD' dataset. A comparative analysis of the performance metrics is presented in Table \ref{table:performance_analysis_of_baseline_models}, which includes accuracy, parameter count (in millions), and model size (in MB). Given our focus on lightweight models suitable for low-end devices, minimizing both the parameter count and model size is crucial.

Among traditional CNN architectures, MobileNetV2 \cite{sandler2019mobilenetv2invertedresidualslinear} emerged as the most efficient in terms of trainable parameters and model size, albeit with a lower accuracy of 91.23\%. In contrast, AlexNet \cite{NIPS2012_c399862d} achieved a higher accuracy but was not optimal for mobile or edge devices due to its substantial model size of 466.6 MB and a parameter count of 58.32 million. DenseNet121 \cite{huang2018denselyconnectedconvolutionalnetworks} offered a more balanced trade-off between accuracy, model size, and parameter count.


In our experiments, the three Vision Transformer models exhibited significantly lower parameter counts while achieving impressive accuracies. EfficientViT\_b0 achieved an accuracy of 98.80\% with only 2.14 million parameters and a model size of 2.04 MB. MobileViT\_s outperformed the others, attaining 99.43\% accuracy with 4.94 million parameters and a model size of 4.71 MB. TinyViT\_5m also demonstrated commendable performance as a lightweight model, 99.10\% accuracy.


\begin{table}[t]
\centering
\caption{Ablation Study}
\label{table:ablation_study}
\begin{tabular}{l c c c}
\toprule
\textbf{Model} & \textbf{Pretrained} & \textbf{Augmentation} & \textbf{Accuracy (\%)} \\ 
\midrule
EfficientViT\_b0 & \ding{55} & \ding{55}   &  67.53 \\ 
EfficientViT\_b0 & \checkmark & \ding{55}   & 98.52 \\ 
EfficientViT\_b0 & \checkmark & \checkmark   & 98.80 \\ 
\midrule
TinyViT\_5m & \ding{55} & \ding{55}  & 75.93 \\ 
TinyViT\_5m & \checkmark & \ding{55}  & 98.73 \\ 
TinyViT\_5m & \checkmark & \checkmark  & 99.10 \\ 
\midrule 
MobileViT\_s & \ding{55} & \ding{55}  & 72.15 \\ 
MobileViT\_s & \checkmark & \ding{55}  & 99.00 \\ 
MobileViT\_s & \checkmark & \checkmark  & \textbf{99.43} \\ 
\bottomrule
\end{tabular}
\end{table}

\subsection{Ablation Study}
An ablation study was conducted to gain a comprehensive understanding of the various components within our experimental framework. This investigation focused on three critical aspects: the choice of the model (EfficientViT\_b0, TinyViT\_5m, and MobileViT\_s), the training status of the models (pretrained versus non-pretrained), and the impact of data augmentation techniques.
To assess the influence of transfer learning on our models, we utilized Vision Transformer architectures pretrained on the ImageNet-1k dataset. The results presented in Table \ref{table:ablation_study} indicate that transfer learning significantly enhances the performance of our Vision Transformer models for the new classification task. Given the superior results achieved with pretrained models, we did not pursue further experiments with non-pretrained versions for other components.

Additionally, the implementation of data augmentation techniques was found to positively impact the accuracy of each Vision Transformer model. By increasing the variability of the training images, the models were able to learn distinct patterns across the various classes more effectively. The combined utilization of pretrained weights and augmentation strategies yielded optimal performance across all models evaluated.
Among the architectures tested, MobileViT\_s achieved the highest accuracy, recording a remarkable 99.43\%.


\subsection{Class-wise Analysis}
\begin{figure}[b]
\centering
\includegraphics[width=0.5\textwidth]{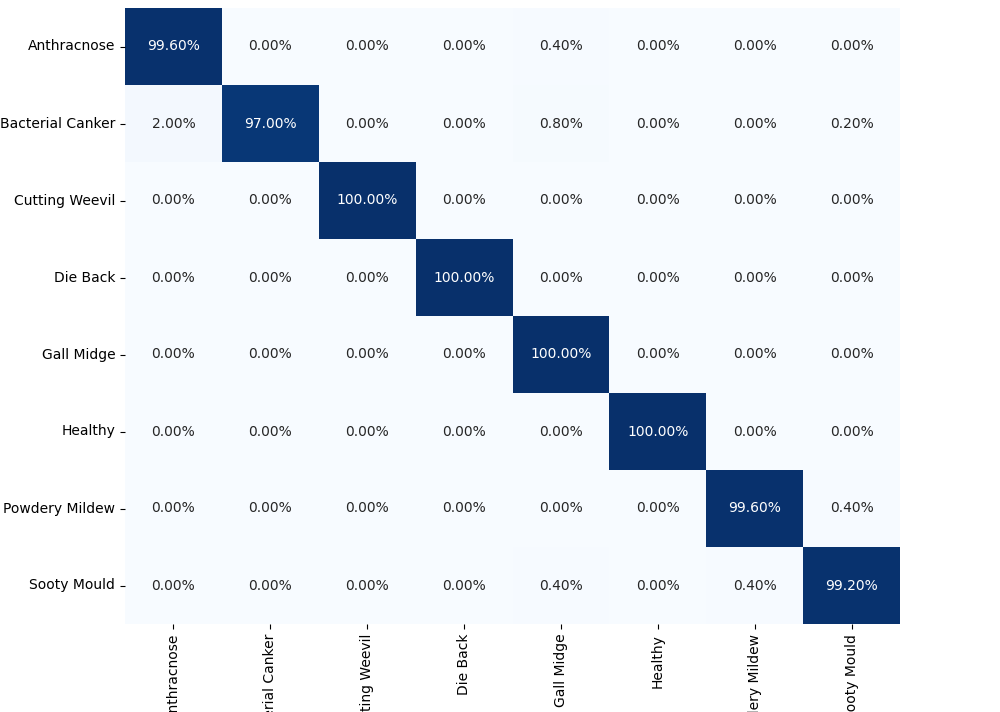}
\caption{Confusion Matrix of MobileViT\_s}
\label{fig:confusion_matrix}
\end{figure}

Table \ref{table:classwise_classification_report} presents a comprehensive summary of the class-wise performance metrics for our optimal model, MobileViT\_s. Notably, the model achieved a recall of 100\% for six out of eight classes, with the exceptions being `Bacterial Canker' and `Sooty Mould', which recorded recall values of 97\% and 99\%, respectively. The slightly lower recall for `Bacterial Canker' can be attributed to inter-class similarities with `Anthracnose', which may have impacted the model's classification accuracy. 
Overall, the average Recall, Precision, and F1-Score across all classes were 99\%, underscoring the model's ability to learn and generalize complex patterns and features by leveraging global contextual information. The confusion matrix, illustrated in Figure \ref{fig:confusion_matrix}, corroborates that the model's performance aligns with expectations for the test split, further validating its efficacy in distinguishing between the various classes.


\begin{table}[t]
\centering
\caption{Class-wise analysis using MobileViT\_s}
\label{table:classwise_classification_report}
\begin{tabular}{l c c c}
\toprule
\textbf{Class} & \textbf{Precision (\%)} & \textbf{Recall (\%)} & \textbf{F1-Score (\%)} \\ 
\midrule
Anthracnose       & 98 & 100 & 99 \\ 
Bacterial Canker  & 100 & 97 & 98 \\ 
Cutting Weevil    & 100 & 100 & 100 \\ 
Die Back          & 100 & 100 & 100 \\ 
Gall Midge        & 98 & 100 & 99 \\ 
Healthy           & 100 & 100 & 100 \\ 
Powdery Mildew    & 100 & 100 & 100 \\ 
Sooty Mould       & 99 & 99 & 99 \\ 
\midrule
\textbf{Average} & \textbf{99} & \textbf{99} & \textbf{99} \\ \bottomrule
\end{tabular}
\end{table}

\subsection{Comparison with state-of-the-art methods}
Table \ref{table:performance_analysis_with_sota_models} presents a comparative analysis of our Vision Transformer (ViT) models against existing state-of-the-art architectures. Our model, MobileViT\_s, achieved an impressive accuracy of 99.43\% while maintaining a lower model size and parameter count. Specifically, EfficientViT\_b0 comprises only 2.14 million trainable parameters and has a model size of 2.04 MB, achieving an accuracy of 98.80\%.

In contrast, the model proposed by Varma \textit{et al.} \cite{Varma2024} attained the highest accuracy of 99.87\%; however, it is significantly larger, with a model size of 20.79 MB and 20.58 million trainable parameters. Additionally, Rizvee \textit{et al.} \cite{RIZVEE2023100787} reported a commendable accuracy of 99.55\% with a model size of 13.1 MB and a higher FLOPs count of 2.26 GFLOPS. In comparison, our EfficientViT\_b0 operates at only 0.1 GFLOPS, while MobileViT\_s maintains a FLOPs count of 1.44 GFLOPS.

This analysis demonstrates that our models not only excel in terms of accuracy but also offer significant efficiency advantages concerning model size, parameter count, and computational complexity, reinforcing their suitability for deployment in resource-constrained environments.


\begin{table}[t]
\centering
\caption{Comparison with state-of-the-art methods on MangoLeafBD Dataset}
\label{table:performance_analysis_with_sota_models}
\begin{tabular}{l p{1.0cm} p{1.5cm} p{1.0cm} p{1.0cm}}
\toprule
\textbf{Architecture} & \textbf{Accuracy (\%)} & \textbf{Trainable Parameters Count (M)} & \textbf{Model Size (MB)} & \textbf{FLOPs Count (GFLOPS)} \\ 
\midrule 
Swapno \textit{et al.}\cite{10585939} & 96.87\% & 120 & 479 & 12.3\\ 
Mahbub \textit{et al.}\cite{10101648} & 98.00\% & -- & -- & --\\ 
Rizvee \textit{et al.}\cite{RIZVEE2023100787} & 99.55\% & 3.26 & 13.1 & 2.26\\ 
Varma \textit{et al.}\cite{Varma2024} & \textbf{99.87\%} & 20.58 & 20.79 & 2.85\\ 

\midrule
\multicolumn{4}{c}{\textbf{Ours}} \\ 
\midrule
EfficientViT\_b0 & 98.80\% & \textbf{2.14} & \textbf{2.04} & \textbf{0.1}\\ 
TinyViT\_5m & 99.10\% & 5.07 & 4.84 & 1.17\\ 
MobileViT\_s & \textbf{99.43\%} & 4.94 & 4.71 & 1.44\\ 
\bottomrule
\end{tabular}
\end{table}


\subsection{Qualitative Analysis}
\begin{figure}[t]
    \centering
    \begin{tabular}{cccccccc}
        \subfloat[Label: Anthracnose]{\includegraphics[width=0.2\textwidth]{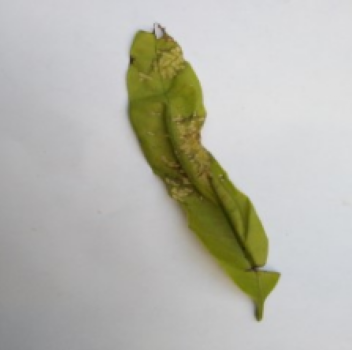}} &
        \subfloat[Predicted: Anthracnose]{\includegraphics[width=0.2\textwidth]{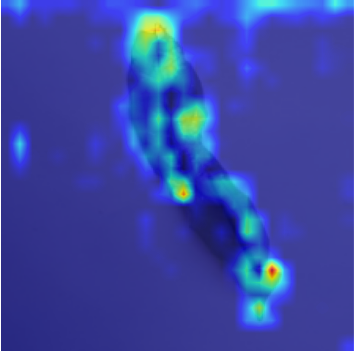}} &
        \\
        \subfloat[Label: Bacterial Canker]{\includegraphics[width=0.2\textwidth]{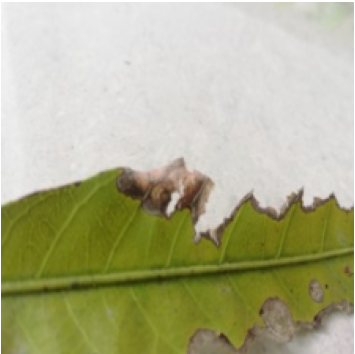}} &
        \subfloat[Predicted: Bacterial Canker]{\includegraphics[width=0.2\textwidth]{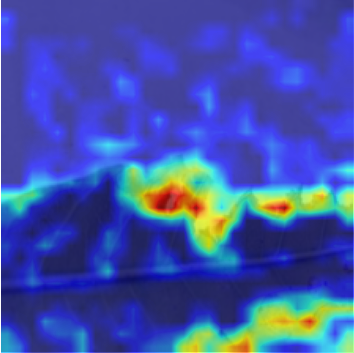}} &
         \\
         \subfloat[Label: Gall Midge]{\includegraphics[width=0.2\textwidth]{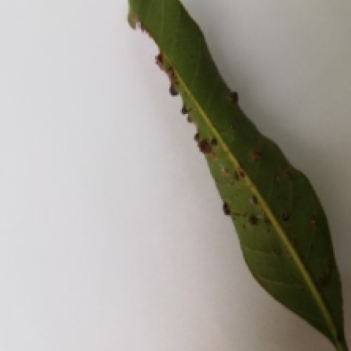}} &
        \subfloat[Predicted: Gall Midge]{\includegraphics[width=0.2\textwidth]{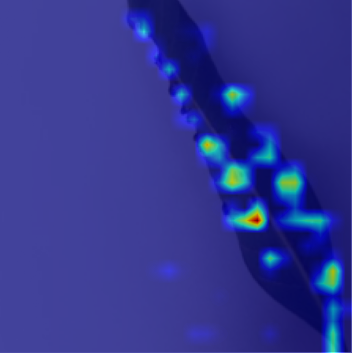}} &
         \\
    \end{tabular}
    \caption{Attention map output of correctly predicted samples.}
    \label{fig:Attention map output of correctly predicted samples}
\end{figure}

To gain an intuitive understanding of our proposed pipeline's ability to accurately classify diseases based on relevant features, we utilized Gradient-weighted Class Activation Mapping (GradCAM) to visualize the model's focus on significant regions in correctly classified samples, as illustrated in Figure \ref{fig:Attention map output of correctly predicted samples} \cite{Selvaraju_2019}. This visualization demonstrates the effectiveness of our model in highlighting specific disease characteristics, showcasing its capability to learn and differentiate unique patterns associated with each class.


Upon analyzing the confusion matrix for our best-performing model, presented in Figure \ref{fig:confusion_matrix}, we observe that the model achieved perfect classification for half of the classes. However, accuracy for the remaining classes was slightly lower, with the majority of misclassifications occurring within the `Bacterial Canker' class. A smaller number of misclassifications were also noted for the `Sooty Mould' and `Powdery Mildew' classes, primarily due to inter-class similarity.

\begin{figure}[h]
    \centering
    \begin{tabular}{cccc}
        \subfloat[Label: Sooty Mould \\ Predicted: Powdery Mildew]{\includegraphics[width=0.2\textwidth,height=3.2cm]{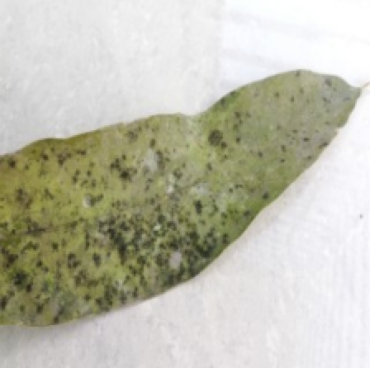}} &
        \hspace{-0.40cm} 
        \subfloat[Label: Powdery Mildew \\ Predicted: Sooty Mould]{\includegraphics[width=0.2\textwidth,height=3.2cm]{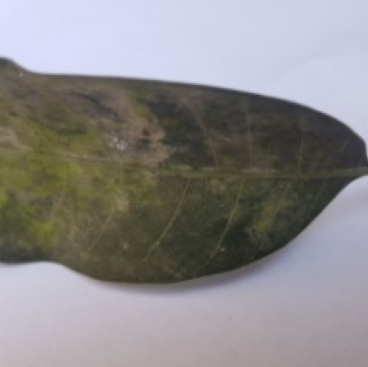}} &
        \hspace{-0.40cm} 
        \\
        \subfloat[Label: Bacterial Canker \\ Predicted: Anthracnose]{\includegraphics[width=0.2\textwidth,height=3.2cm]{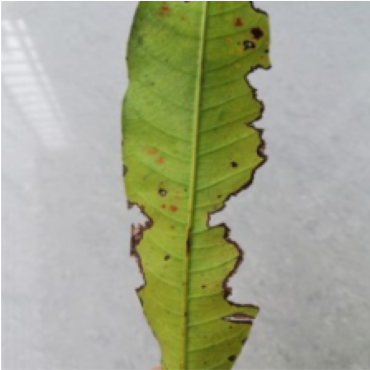}} &
        \hspace{-0.40cm} 
        \subfloat[Label: Anthracnose]{\includegraphics[width=0.2\textwidth, height=3.2cm]{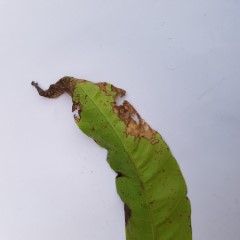}} &
        \hspace{-0.40cm} 
    \end{tabular}
    \caption{Misclassified sample with visually similar samples of the predicted class.}
    \label{fig:inter class similarity samples}
\end{figure}

As shown in Figure \ref{fig:inter class similarity samples}a and Figure \ref{fig:inter class similarity samples}b, the `Sooty Mould' and `Powdery Mildew' classes exhibit similar patterns, which contributed to the model’s tendency to misclassify these instances. Additionally, Figures \ref{fig:inter class similarity samples}c and \ref{fig:inter class similarity samples}d reveal that certain images of `Bacterial Canker' were incorrectly predicted as `Anthracnose', again highlighting the impact of inter-class similarities on classification accuracy. This phenomenon of overlapping features among classes is a significant factor contributing to the model’s misclassification errors.


\section{Conclusion}\label{conclusion}

The efficient and accurate classification of mango leaf diseases is pivotal to meeting the growing demands of mango production. In this work, we proposed a lightweight Vision Transformer architecture with a self-attention mechanism to effectively learn features and patterns, thereby capturing global contextual information from leaf images. Furthermore, we employed augmentation techniques to introduce variability within the classes and leveraged transfer learning from large datasets to enhance model accuracy.
Despite the presence of both inter-class and intra-class variations in our dataset, our experimental results demonstrate that our models outperform state-of-the-art approaches in terms of accuracy, model size, parameter count, and FLOPs. Additionally, we validated the effectiveness of our models through attention map visualizations and provided samples for error analysis.
Future work can focus on expanding this dataset by collecting samples directly from field conditions to enhance real-world applicability. 


\bibliographystyle{ieeetr}
\bibliography{citations}

\end{document}